# How many prompts, models, and runs? A generalizability theory tutorial for large language model ratings of L2 pragmatic appropriateness

Katikela Sreeharsha Kishore
Research Scholar in Applied Linguistics
Centre for Applied Linguistics & Translation Studies, University of Hyderabad

**Abstract**

Scores generated using large language models (LLMs) are affected by prompts, models, and iterations. In validity studies in applied linguistics, high intercoder reliability has often been observed between scores generated with one configuration and those generated by human coders, and/or between repeated scores using the same configuration. This methods paper explains how generalizability theory can be used to determine the amount of variance in scores attributed to different facets (i.e., prompt, model, and iteration) and the number of facets required to obtain reliable scores. As an example, scores of appropriateness in 120 L2 written requests in English were extracted using 7,200 ratings by LLMs and 360 ratings by human coders. Although high inter-rater reliability was found between repeated iterations of the same configuration (Krippendorff's α: .91-.94), generalizability coefficients for scores from a single configuration were relatively low (G=.74) when considering prompts and models as sources of error. The largest source of error was response x model interaction. Adding more iterations had the smallest effect on improving generalizability, whilst spreading ratings across models and prompts had the largest effect. Using 4 prompts and 3 models achieved moderate agreement with scores from trained human coders (r=.88; quadratic weighted κ=.72; standardised mean difference=.01). However, longer responses were rewarded in the scores (β=0.23 points per SD, p<.001). Measures of reliability and agreement did not indicate this systematic bias. The advantages and disadvantages of different approaches are discussed. Considerations in this process include limited levels within facets, model drift, and shared bias across models.

**Keywords:** Large language models; Generalizability theory; LLM-as-a-judge; Rater reliability; Automated scoring; L2 pragmatics

## 1. Introduction

Large language models (LLMs) are increasingly applied as raters in applied linguistics research. LLMs have been used to score L2 writing (Gayed, 2026; Mizumoto & Eguchi, 2023; Yamashita, 2024), evaluate accuracy in writing (Pfau et al., 2023), categorise learner free responses (Mizumoto & Teng, 2025), annotate for rhetorical moves (Grullon-Polanco, 2026; Yu et al., 2026), give psycholinguistic ratings (Zhong et al., 2025) and evaluate pragmatic competence in L2 learners (Mahmoudi-Dehaki & Nasr-Esfahani, 2025). However, ratings given by LLMs are sensitive to differences in configuration (different model, prompt and sampling settings) that could plausibly be made by other researchers.

Typically in studies using LLMs, inter-rater reliability is reported between one configuration and human raters, and increasingly, reliability between multiple runs of the same configuration (Grullon-Polanco, 2026). This does not indicate whether findings would be consistent if a different acceptable prompt/model configuration were used. Evidence from natural language processing indicates that findings may well differ. Small rewordings to prompts can change ratings significantly (Sclar et al., 2024). LLM raters demonstrate bias in ratings (Saito et al., 2023; Wang et al., 2024; Zheng et al., 2023) and in 37 replicated annotation tasks, approximately 31% of hypothesis tests gave incorrect conclusions depending on the configuration used (Baumann et al., 2025). Reviews of AI-assisted assessments in applied linguistics confirm that studies typically focus on inter-rater reliability with just one human reference, with few studies focusing on generalisability across conditions, and very few explicitly considering validity (Han, 2026; Kim et al., 2026). (See also: Barrot, 2026).

Generalisability theory (G-theory) (Brennan, 2001; Cronbach et al., 1972; Shavelson & Webb, 1991) provides a solution. G-theory has been widely used in language testing to partition variance in ratings due to raters, tasks and occasions (Bachman et al., 1995; Lynch & McNamara, 1998). It has also been applied to examine reliability of questionnaires in this journal (Lee & Ye, 2023). Within G-theory, prompt wording, model and number of sampling runs can be considered facets of measurement. By doing so, we can estimate how much variance in ratings is caused by each factor. We can then plan for reliable ratings by determining how many prompts, models and runs are required. While generalisability theory is starting to be applied in LLM evaluation (Li & Lee, 2026; Messing, 2026), it has not yet been presented to applied linguists using LLMs as raters.

This tutorial aims to address this issue. Some knowledge of R is assumed. A description of generalisability theory is given in section 2, an example is described in section 3, different options are compared in section 4, and issues to be aware of are discussed (with a reporting checklist at the end) in section 5. An example from pragmatic competence in L2 learners is used as a case study. Inter-rater reliability is a known issue in pragmatic competence (Taguchi, 2011), and LLMs are now beginning to be used to evaluate pragmatic appropriateness (Mahmoudi-Dehaki & Nasr-Esfahani, 2025). However, this method can be used for any rating/coding task.

## 2. What is it? Generalizability theory for LLM rating designs

Generalizability theory (G-theory) asks how far a score obtained under one set of conditions generalizes to the scores that could have been obtained under other conditions the researcher considers equally acceptable (Brennan, 2001). Applying it to LLM ratings requires three decisions.

**The object of measurement.** This is what the scores are meant to differentiate: here, learner responses (*p*).

**The universe of admissible observations.** These are the rating conditions over which the researcher wants to generalize. In LLM rating, three facets are usually relevant. *Prompts* (*q*) are alternative wordings of the rating instruction that convey the same rubric, scale, and descriptors. *Models* (*m*) are the LLMs the researcher would accept as raters. *Runs* (*r*) are repeated, independent calls under identical inputs, which differ because outputs are sampled stochastically. Defining what counts as an admissible prompt is a substantive decision, not a technical one. A paraphrase that changes the construct (e.g., adding a criterion) is a different instrument, not a different prompt.

**The design.** If every response is rated under every prompt–model combination, and each combination is called several times, the design is responses crossed with prompts and models, with runs nested within each cell: $r{:}(p \times q \times m)$. The observed rating is then decomposed into a grand mean plus effects for each facet and interaction (Eq. 1):

$$X_{pqmr} = \mu + \nu_p + \nu_q + \nu_m + \nu_{pq} + \nu_{pm} + \nu_{qm} + \nu_{pqm} + \nu_{r:pqm,e} \qquad (1)$$

Each effect has a variance component with a concrete interpretation. The component for *p* is universe-score variance, the signal. The components for *q* and *m* index systematic severity differences between prompts and between models. The *pq* and *pm* components index how far prompts and models *reorder* responses. The *qm* and *pqm* components are higher-order interactions. The residual, *r:pqm*,*e*, captures run-to-run sampling variation confounded with unexplained error.

A *G-study* estimates these components. A *D-study* then uses them to project the dependability of scores averaged over *n'q* prompts, *n'm* models, and *n'r* runs. Relative error variance (Eq. 2) contains the components that change the rank order of responses; absolute error variance (Eq. 3) adds the components that shift all scores up or down:

$$\sigma^2(\delta) = \frac{\sigma^2_{pq}}{n'_q} + \frac{\sigma^2_{pm}}{n'_m} + \frac{\sigma^2_{pqm}}{n'_q n'_m} + \frac{\sigma^2_{r:pqm,e}}{n'_q n'_m n'_r} \qquad (2)$$

$$\sigma^2(\Delta) = \sigma^2(\delta) + \frac{\sigma^2_{q}}{n'_q} + \frac{\sigma^2_{m}}{n'_m} + \frac{\sigma^2_{qm}}{n'_q n'_m} \qquad (3)$$

The generalizability coefficient, $E\rho^2 = \sigma^2 p / [\sigma^2 p + \sigma^2(\delta)]$, is appropriate when scores are used to rank or correlate (e.g., relating pragmatic appropriateness to proficiency). The dependability coefficient, $\Phi = \sigma^2 p / [\sigma^2 p + \sigma^2(\Delta)]$, is appropriate when absolute score levels matter (e.g., cut scores, comparisons with human means, or comparisons across studies that used different prompts).

The generalization of this study can be interpreted via this framework. Within a given prompt and model pair, pq, pm and pqm effects will be consistent across runs. As such, inter-run reliability provides generalization to a universe containing only this prompt and model.

Considerations: Firstly, prompts and runs may be random factors, but models may be random or fixed. If general inferences are to be made about competent LLMs, models should be random. However, if a specific set of models is being used, models should be fixed. In this case, the pm effect (divided by the number of models) moves from error to variability in universe scores (Brennan, 2001). Secondly, generalization does not equate to validity. If configurations have been trained on overlapping data, they may agree due to sharing similar flaws (Gerlach, 2026). If a construct irrelevant property is rewarded across all configurations, such as the length of responses, then $\sigma^2 p$ will increase, and so will $E\rho^2$. If using an argument-based framework (Kane, 2013), a G study only provides evidence for generalization. Generalization to the intended construct requires additional evidence, typically comparison between the ratings of trained human raters, and targeted investigations for bias (Williamson et al., 2012). This is provided in steps 5 and 6 below.

## 3. How is it done? A worked example

### *3.1. Data*

The data are artificially generated. In this example, 120 L2 English learners wrote a written discourse completion task requesting an extension of time from a professor. Each response was scored using a 6-point holistic scoring scheme (0-5), as used by Taguchi (2006). Three human raters scored each response. In addition, each response was scored by three LLMs (M1-M3) using four versions of the same scoring rubric (P1-P4). Each LLM scored each response 5 times in an independent run, resulting in 7200 LLM scores. The four versions of the rubric differ only in irrelevant construct-irrelevant aspects, including the order of descriptors, role descriptions and requested output type.

Scores were generated according to Equation 1 with known variances within the model, and rounded and truncated to be within the appropriate range (0 to 5). LLM scores were inflated by 0.20 so that the LLM mean equaled the human mean. This allows the resulting composite to meet traditional inter-scorer agreement thresholds (Step 5). A construct irrelevant variance source was also introduced. For example, LLM scores, but not human scores, increase with longer responses. This can be detected as described in Step 6. The values given here are for illustrative purposes only, and do not represent estimated values for any model.

### *3.2. Step 1: The usual check*

Most studies stop at the consistency of a single configuration across runs. Here, Krippendorff's α (interval) across the five runs ranged from .91 to .94 (median = .93) for the 12 prompt–model configurations. By conventional standards (Krippendorff, 2019), every configuration looks highly reliable. Steps 2 and 3 show why that conclusion does not follow.

### *3.3. Step 2: Estimate the variance components (G-study)*

The components are estimated with a linear mixed model in which every facet and interaction is a random intercept (Jiang, 2018), fitted with the lme4 package (Bates et al., 2015):

```
g_fit <- lmer(score ~ 1 + (1|resp) + (1|prompt) + (1|model) +
          (1|resp:prompt) + (1|resp:model) + (1|prompt:model) +
          (1|resp:prompt:model), data = llm, REML = TRUE)
```

The residual variance is the *r:pqm,e* component. For balanced designs, the classical ANOVA estimator (Brennan, 2001) gives the same values; here they were identical to three decimals. The mixed-model approach also tolerates missing cells, such as unparsable outputs: deleting 3% of ratings at random changed no component by more than .001.

**Table 1.** G-study variance components for LLM ratings (simulated data).

| Source | Interpretation | $\sigma^2$ | % | 95% CI |
|---|---|---|---|---|

| Source | Interpretation | $\sigma^2$ | % | 95% CI |
|---|---|---|---|---|
| Responses (*p*) | Universe score (signal) | 1.089 | 67.5 | [0.845, 1.456] |
| Prompts (*q*) | Prompt severity | 0.053 | 3.3 | [0.016, 0.996] |
| Models (*m*) | Model severity | 0.077 | 4.8 | [0.020, 4.136] |
| $p \times q$ | Prompts reorder responses | 0.052 | 3.2 | [0.041, 0.068] |
| $p \times m$ | Models reorder responses | 0.154 | 9.5 | [0.126, 0.191] |
| $q \times m$ | Prompt severity differs by model | 0.009 | 0.5 | [0.003, 0.052] |
| $p \times q \times m$ | Three-way interaction | 0.072 | 4.5 | [0.064, 0.083] |
| *r*:*pqm*, *e* | Run-to-run variation, residual | 0.108 | 6.7 | [0.104, 0.112] |

*Note.* $\sigma^2$ in squared scale points. Confidence intervals are Satterthwaite intervals (Brennan, 2001).

Responses account for 67.5% of the variance. The largest error source, however, is the response × model interaction (9.5%): the three models disagree about which responses are more appropriate. Run-to-run variation (6.7%) is the only error source that an inter-run check can detect. Severity differences are also present: mean ratings ranged from 2.41 to 2.97 across models and from 2.47 to 3.04 across prompts. The confidence intervals for prompts and models are very wide, because a variance estimated from four or three levels is imprecise (Section 5).

### *3.4. Step 3: Plan the design (D-study)*

Table 2 applies Eqs. 2 and 3 to the Table 1 estimates.

**Table 2.** Projected generalizability ($E\rho^2$) and dependability ($\Phi$) by design.

| Design (prompts × models × runs) | Calls per response | $E\rho^2$ | $\Phi$ |
|---|---|---|---|
| 1 × 1 × 1 | 1 | .74 | .67 |
| 1 × 1 × 5 | 5 | .78 | .71 |
| 4 × 1 × 1 | 4 | .84 | .78 |
| 1 × 3 × 1 | 3 | .87 | .82 |
| 2 × 2 × 1 | 4 | .88 | .84 |
| 2 × 3 × 1 | 6 | .91 | .87 |
| 4 × 3 × 1 | 12 | .93 | .90 |
| 4 × 3 × 5 | 60 | .94 | .91 |

Three lessons follow. First, a single configuration that looked excellent in Step 1 ($\alpha$ = .93) has $E\rho^2$ = .74 and $\Phi$ = .67 once prompts and models are treated as sources of error. The inter-run figure equals what Table 1 implies when the *pq*, *pm*, and *pqm* components are counted as signal (.93), as Section 2 predicted. Second, repeating runs is the least efficient investment: five runs of one configuration raise $E\rho^2$ only from .74 to .78, at five times the cost. Third, spreading a fixed budget across models and prompts pays most. Four calls arranged as two prompts × two models give .88, whereas four prompts with one model give .84. Fig. 1 shows the full pattern. For these data, the reliability that Step 1 seemed to promise (.93) is reached only when about three prompt wordings are averaged across three models.

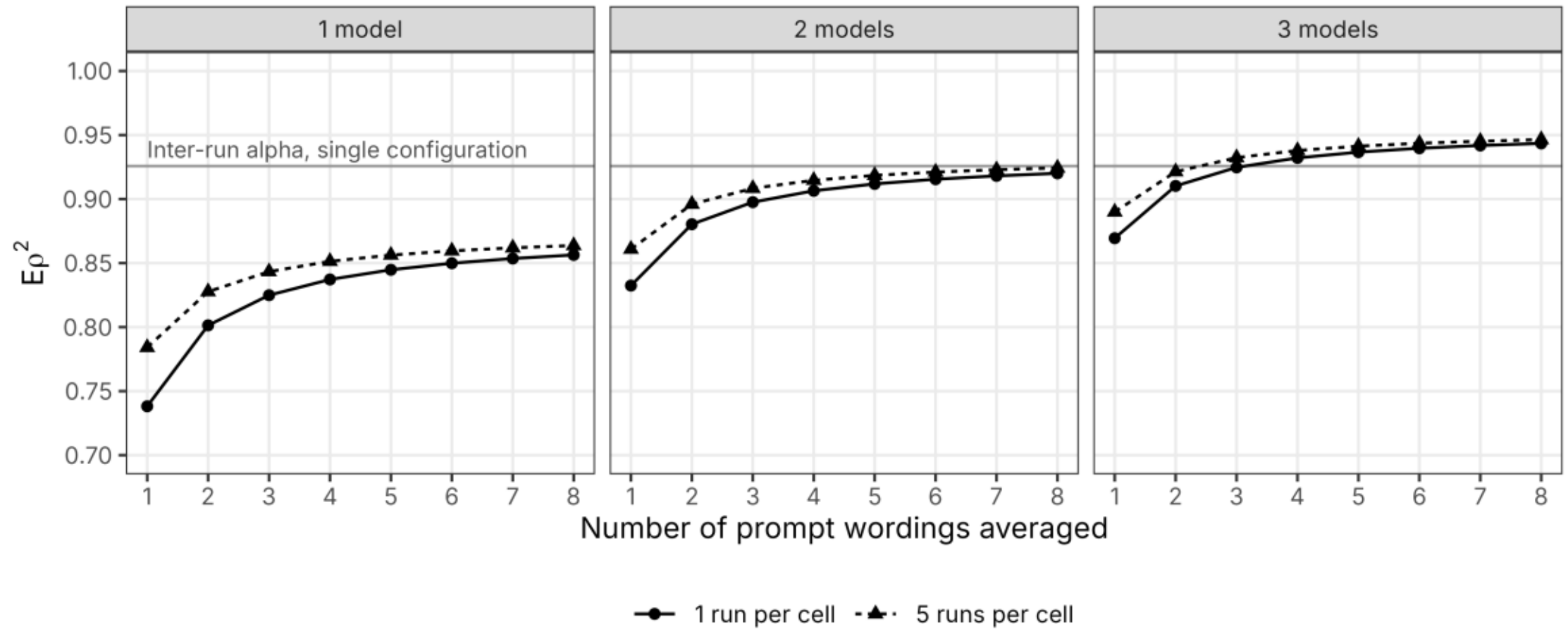

**Fig. 1.** Projected generalizability coefficients ($E\rho^2$) by number of prompt wordings, models, and runs. The horizontal line marks the median inter-run Krippendorff's α of a single prompt–model configuration (.93). Simulated data.

### *3.5. Step 4: Decide whether models are fixed or random*

If the three models are themselves the instrument, and no generalization to other models is intended, the model facet is fixed. For the 4 × 3 × 1 design, $E\rho^2$ then rises from .93 to .98 and Φ from .90 to .96, because the response × model interaction becomes part of the universe score. The higher figures are legitimate only for the narrower claim: the scores describe what this panel of models judges, not what any capable LLM rater would judge. Reports should state which claim is made.

### *3.6. Step 5: Gather criterion evidence against human raters*

A parallel G-study of the human ratings (responses × raters) gave $\sigma^2 p = 0.871$, $\sigma^2 h = 0.145$, and $\sigma^2 ph,e = 0.391$, so $E\rho^2 = .69$ for one rater and .87 for the mean of three. The LLM composite from the 4 × 3 × 1 design was then compared with the human ratings using the indices recommended by Williamson et al. (2012).

**Table 3.** Criterion evidence for the LLM composite (4 prompts × 3 models × 1 run).

| Index | Value | Common benchmark |
|---|---|---|
| Pearson *r*, LLM composite vs human mean | .88 | ≥ .70 |
| *r* corrected for unreliability of both scores | .98 | — |
| QWK, rounded LLM composite vs each human rater (mean) | .72 | ≥ .70 |
| QWK, human vs human (mean of three pairs) | .62 | — |

| Index | Value | Common benchmark |
|---|---|---|
| Degradation (human–human minus LLM–human QWK) | −.09 | ≤ .10 |
| Standardized mean difference (LLM − human) | .01 | \|SMD\| ≤ .15 |

*Note.* QWK = quadratic weighted kappa (Cohen, 1968). Benchmarks follow Williamson et al. (2012).

By every conventional criterion, the LLM composite agrees with trained raters at least as well as they agree with one another.

### *3.7. Step 6: Probe for construct-irrelevant variance*

High generalizability and high agreement do not rule out systematic bias. LLM judges tend to favour longer texts (Saito et al., 2023; Zheng et al., 2023), and in request production, length is partly confounded with appropriateness because supportive moves add words. The probe therefore asks whether LLM scores reward length *beyond* what human raters do. The difference between the LLM composite and the human mean was regressed on standardized response length ($M$ = 31.4 words, $SD$ = 9.5). The slope was 0.23 scale points per $SD$ of length, 95% CI [0.15, 0.32], $p < .001$, $R^2 = .20$ (Fig. 2). Length correlated .38 with human scores but .56 with LLM scores. Consistent with this, LLM universe-score variance (1.089) exceeded human universe-score variance (0.871): part of the "signal" in Table 1 is length.

This is the built-in effect, and none of Steps 1–5 detected it. Any substantive analysis using these scores would have favoured variables correlated with length, such as proficiency. Remedies include adding a length-neutral instruction to every prompt, calibrating LLM scores against a human-rated subsample, or controlling for length in substantive analyses. Each remedy should be followed by repeating Steps 2–6.

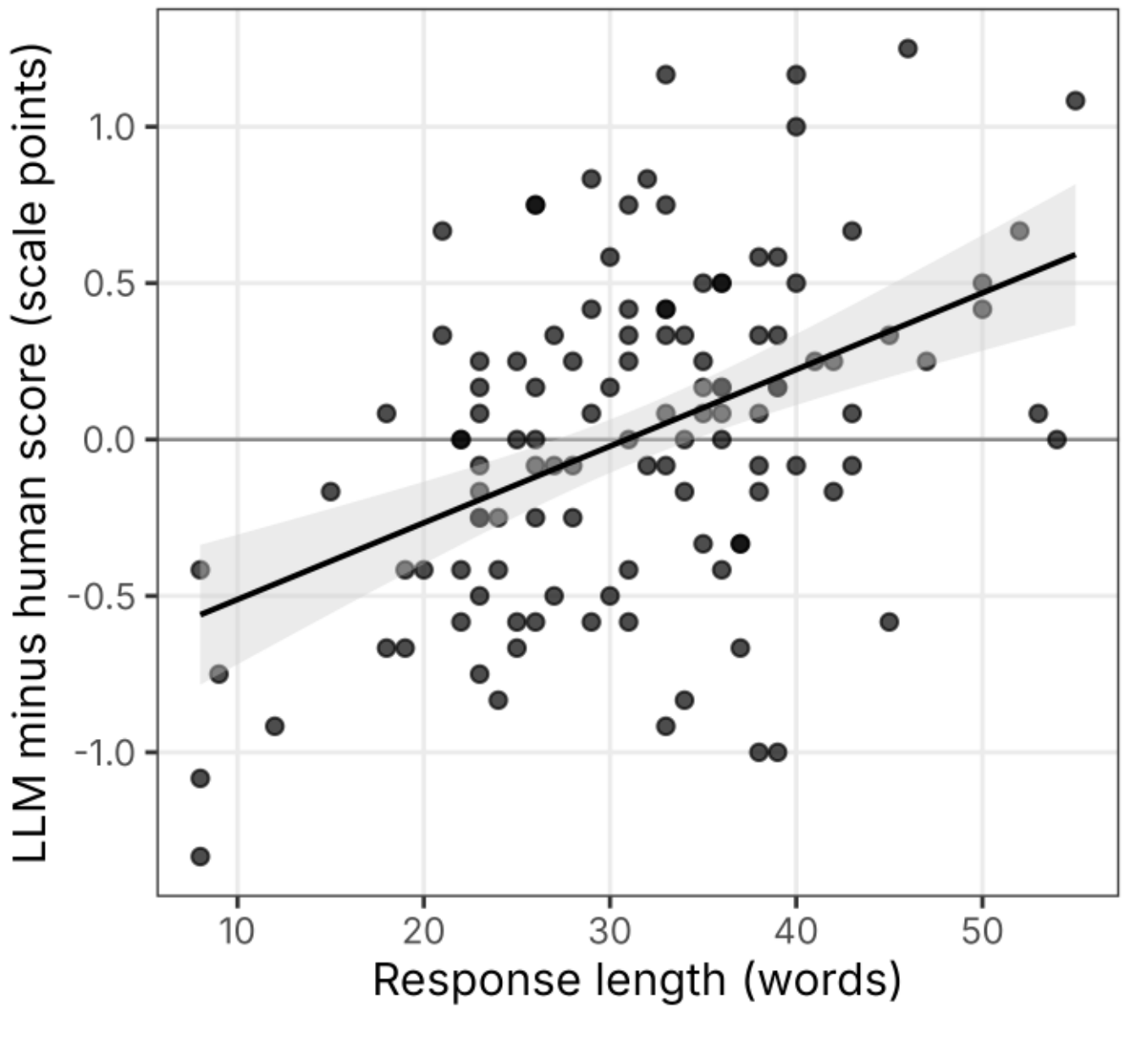

**Fig. 2.** Difference between the LLM composite and the human mean as a function of response length, with ordinary least squares fit and 95% confidence band. Simulated data.

## 4. What options are there?

G-theory is one of several ways to evaluate LLM ratings, and the options answer different questions (Table 4). They are complements rather than rivals.

**Table 4.** Options for evaluating LLM ratings.

| Approach | Question answered | Strengths | Limitations |
|---|---|---|---|
| Percent agreement, Cohen's κ, QWK | Do two sources assign the same scores? | Familiar; benchmarks exist (Williamson et al., 2012) | Pairwise; one condition at a time |
| Intraclass correlation | How consistent are several raters? | Several raters; interval scores | One facet at a time; several forms to choose between |
| Krippendorff's α | How far do coders agree beyond chance? | Any number of coders; missing data; any metric (Krippendorff, 2019) | One facet at a time; does not partition error |
| Many-facet Rasch measurement | How severe or inconsistent is each rater, and where is there bias? | Adjusts for severity; flags individual configurations | Group-level dependability not directly projected; near-deterministic raters overfit (Yamashita, 2024) |
| Generalizability theory | How much does each facet contribute, and how many conditions are needed? | Several facets at once; D-study planning | Group-level; assumes interval scores; imprecise with few facet levels |
| Mixed-effects or Bayesian ordinal models | Same partition, with ordinal scales and covariates | Respects the scale; tests bias directly; full uncertainty | More complex; reliability must be derived |

In language testing, G-theory and many-facet Rasch measurement have long been combined: the first for designing the measurement, the second for diagnosing individual raters (Bachman et al., 1995; Lynch & McNamara, 1998). The same division of labour suits LLM rating. For most research uses, a defensible minimum is a G-study and D-study to justify the design, agreement indices against trained human raters for criterion evidence, and regression probes for the construct-irrelevant features most likely to matter. Many-facet Rasch measurement is added when individual prompts or models need to be flagged and their scores adjusted.

## 5. What issues might there be, and how can they be resolved?

**Too few levels of a facet.** The intervals in Table 1 show that variances estimated from three models or four prompts are highly uncertain. Prompt wordings are cheap to add: six to ten vetted paraphrases make $\sigma^2 q$ far more stable. Because the number of usable models is limited, researchers should report intervals, consider Bayesian estimation, or treat models as fixed and restrict their claims accordingly.

**An ill-defined universe of prompts.** Paraphrases that alter the rubric inflate error, and near-identical paraphrases conceal it. The rubric should be fixed first. Wordings should then vary features that are irrelevant to the construct, such as descriptor order, role framing, and output format (Sclar et al., 2024), and two experts should confirm their equivalence before data collection.

**Relying on low temperature.** Low sampling temperature improves inter-run consistency (Grullon-Polanco, 2026), but it does nothing about prompt or model variance. Highly consistent LLM ratings can also be hard to analyse with measurement models that expect rater variability (Yamashita, 2024). Sampling settings should be reported, and prompts and models should still be sampled.

**Ordinal scores and boundary effects.** G-theory treats ratings as interval data. With short scales and many scores at the floor or ceiling, results should be checked against an ordinal mixed model and the score distribution reported.

**Missing and unparsable output.** Malformed outputs are common at some settings (Grullon-Polanco, 2026). Their rate should be reported, and variance components should be estimated with REML, which tolerates unbalanced data (Section 3.3).

**Model drift.** Proprietary models are updated without notice. Researchers should record version identifiers and access dates, archive raw outputs, and prefer open-weight models where feasible (Grullon-Polanco, 2026).

**Agreement mistaken for validity.** Models trained on overlapping data can share blind spots, so consistency across models is not evidence of accuracy (Gerlach, 2026). A human-rated subsample and probes for likely construct-irrelevant features (length, spelling, L1-related forms, formulaic politeness markers) are therefore essential. The difference-score probe in Step 6 assumes that LLM and human scores share a unit. If LLM scores are more spread out, the difference will correlate with anything related to ability; regressing the difference on the human mean as well, or using a latent-variable model, guards against this.

**Learner data and third-party services.** Consent forms should cover automated processing, and texts sent to commercial services should be de-identified; locally run open-weight models avoid the issue.

Table 5 summarizes what a study using LLM ratings should report.

**Table 5.** Reporting checklist for studies using LLM ratings.

| Area | What to report |
|---|---|
| Construct and rubric | Full rubric and scale; human rater training |

| Area | What to report |
|---|---|
| Prompts | All wordings; how equivalence was established |
| Models | Names, version identifiers, access dates, open or closed weights |
| Sampling | Temperature and other parameters; number of runs |
| Output handling | Parsing rules; rate and treatment of unusable outputs |
| G-study | Design; estimator; all variance components with intervals |
| D-study | Final design; $E\rho^2$ and $\Phi$; which facets are fixed, and why |
| Criterion evidence | Human G-study; *r*, QWK, degradation, and SMD against human raters |
| Bias probes | Features tested; results; remedies applied |
| Materials | Prompts, raw outputs, and code in an open repository such as IRIS or OSF |

## 6. Conclusion

LLM ratings are measurements, and their conditions of measurement are facets. Reporting the run-to-run consistency of one prompt–model configuration overstates how dependable the scores are. In the worked example, a configuration with an inter-run α of .93 had a generalizability coefficient of .74 once prompts and models were treated as sources of error. Extra runs bought little, while spreading calls across models and prompts bought most. The example also showed that a composite meeting every conventional agreement benchmark could still reward response length. G-theory makes the first problem visible and allows the design to be planned. Only human anchoring and targeted probes reveal the second. Together, they let applied linguists replace the question "Does the LLM agree with humans?" with two more useful ones: over which prompts, models, and runs do these scores generalize, and what else do they measure?

## Data availability

All data reported in this article are simulated. The R script that generates and analyses them, the resulting data files , and example prompt wordings, will be available on request.

## Declaration of generative AI and AI-assisted technologies in the manuscript preparation process

During the preparation of this work, the author(s) did not use any AI tools. The author reviewed and edited the content as needed and takes full responsibility for the content of the published article.